\documentclass[cameraready]{Interspeech}

\usepackage{amsmath}
\usepackage{cite}
\usepackage{multirow}
\usepackage{CJKutf8}

\title{Benchmarking Large Language Models for Grapheme-to-Phoneme Conversion: 
A Japanese Case Study}

\author[affiliation={1}, orcid=0000-0002-8347-5604]{Tomoki}{Koriyama}
\address{$^1$ CyberAgent, Japan}
\email{koriyama\_tomoki@cyberagent.co.jp}
\keywords{grapheme-to-phoneme, large language model, text-to-speech, benchmark}

\begin{document}
\maketitle

\begin{abstract}
Grapheme-to-phoneme (G2P) conversion is essential for controllable and robust text-to-speech, and large language models (LLMs), with broad linguistic knowledge, offer a promising approach.
We benchmarked over 30 LLMs on Japanese G2P, comparing them with conventional morphological analyzers on 3000 manually annotated sentences.
We evaluated two prompting strategies: a parse mode, where the LLM performs morphological analysis followed by rule-based kana conversion, and a direct mode, where the LLM directly predicts kana readings.
The results show that model size, version, and Japanese-specialized training are key factors, with the best LLMs achieving kana character error rate below 0.52\% vs. the best conventional tool (1.03\%).
Parse mode outperforms direct mode for most models, as rule-based post-processing relieves the LLM of handling complex pronunciation rules.
We also show that feeding LLM-predicted kana into a kana-input TTS yields better pronunciation than end-to-end TTS.
\end{abstract}

\section{Introduction}

Grapheme-to-phoneme (G2P) conversion is a core component of text-to-speech (TTS) systems,
in which written text is transformed into phonetic representations~\cite{chen2003conditional, yao2015seq2seq, deri2016g2p, mortensen2018epitran, ashby2021sigmorphon}.
Recent end-to-end (E2E) TTS including large language model (LLM)-based ones directly convert text to speech waveforms~\cite{kim2021conditional, wang2023valle, du2024cosyvoice2, chen2024f5tts},
implicitly learning pronunciation rules from training data
and reducing the need for an explicit G2P module.
Despite this trend, explicit G2P remains important for practical TTS applications for controllability and robustness.
For example, users need to control pronunciation,
particularly for proper nouns such as people's names and place names.
Moreover, E2E models often produce incorrect pronunciation that sounds very unnatural for listeners.

G2P is a challenging task in many languages
because the mapping from graphemes to phonemes is often context-dependent.
Heteronyms, such as \textit{read} (/\textipa{{\*r}i:d}/ vs. /\textipa{{\*r}Ed}/) in English,
require disambiguation beyond simple dictionary lookup.
Japanese G2P presents additional unique challenges.
Japanese text is written without explicit word boundaries, requiring morphological analysis.
Kanji characters used in Japanese are frequently polyphonic, having multiple readings depending on context
(e.g., \begin{CJK}{UTF8}{ipxm}今\end{CJK} can be \textit{ima} or \textit{kon}).
Furthermore, numeral expressions combined with counters have irregular readings
(e.g., \begin{CJK}{UTF8}{ipxm}1本\end{CJK} is \textit{ippon}, not \textit{ichihon}).
These challenges make simple dictionary lookup insufficient for general use.
Rule-based systems such as OpenJTalk~\cite{openjtalk} and MeCab with the UniDic dictionary~\cite{kudo2004mecab}
have been developed and are widely used in Japanese G2P.
Statistical and DNN-based approaches have also been explored~\cite{hatori2011japanese, kakegawa21phonetic, kurihara24japanese, shirahata2026ccg}.

LLMs can be used as an alternative to conventional G2P tools
because their pretraining knowledge eliminates the need for task-specific training data
and enables handling of out-of-vocabulary words and loanwords.
In addition, LLMs support in-context learning for specifying pronunciation via prompting.
With these capabilities, LLMs have advanced morphological segmentation and part-of-speech tagging~\cite{zhao2023survey, batsuren2024llmsegm, machado2024pos, fang2025chinese}
and text normalization~\cite{ma2026textnorm}.
For G2P specifically, Suvarna et al.~\cite{suvarna2024phonologybench} evaluated the phonological skills of LLMs for English,
showing that recent LLMs achieve reasonable G2P accuracy.
Han et al.~\cite{han2024improving} improved English G2P
by using in-context knowledge retrieval with GPT-4.
Fetrat Qharabagh et al.~\cite{mahta2024llm} benchmarked LLM-based G2P for Persian,
showing that a cascade approach---first transliterating Persian script into Latin script (Finglish) using an LLM, then converting to phonemes---is effective.

In this study, we evaluate two approaches for LLM-based Japanese G2P: \textit{parse} and \textit{direct} modes.
In parse mode, the LLM performs morphological analysis and estimates the reading of each word,
followed by rule-based post-processing for pronunciation rules.
In direct mode, the LLM directly predicts the kana reading of the entire text.
We evaluate a wide range of LLMs
varying in model size, version, and Japanese-specialized training,
including both proprietary API models and locally deployed open-weight models.
G2P accuracy is measured by kana character error rate (CER).
The experimental results show that larger models are particularly effective for G2P,
that parse mode outperforms direct mode for most models,
and that Japanese-specialized training greatly improves accuracy.

Our main contributions are as follows:
(1)~We present the first large-scale benchmark of LLM-based Japanese G2P,
covering over 30 models across proprietary APIs, open-weight LLMs, and conventional morphological analyzers.
(2)~We show that LLM-based G2P combined with a kana-input TTS system
achieves better pronunciation accuracy than E2E TTS systems while maintaining comparable naturalness.
(3)~We release manual kana reading annotations for 3,000 sentences\footnote{The dataset and evaluation script are available at \url{https://github.com/CyberAgentAILab/jvs_nonpara_kana}}.

\section{Japanese Grapheme-to-Phoneme Conversion}

Japanese grapheme-to-phoneme requires converting mixed-script text
(kanji, hiragana, katakana, numerals, and Latin characters)
into a phonetic representation.
Accurate G2P conversion requires consideration of multiple linguistic factors.

\textbf{Word segmentation.}
Japanese text is written without spaces between words,
requiring morphological analysis to identify word boundaries and parts of speech.
Incorrect segmentation directly causes reading errors;
for example, \begin{CJK}{UTF8}{ipxm}米原発\end{CJK} can be segmented as
\begin{CJK}{UTF8}{ipxm}米\end{CJK}+\begin{CJK}{UTF8}{ipxm}原発\end{CJK} (\textit{bei-genpatsu}, American nuclear power plant)
or \begin{CJK}{UTF8}{ipxm}米原\end{CJK}+\begin{CJK}{UTF8}{ipxm}発\end{CJK} (\textit{maibara-hatsu}, departing from Maibara)
depending on its context.

\textbf{Reading estimation.}
Each word must be assigned a kana reading.
This is straightforward for kana-only words but challenging for kanji compounds.
Polyphonic kanji such as \begin{CJK}{UTF8}{ipxm}行\end{CJK}
(\textit{i}/\textit{yu}/\textit{koo}/\textit{gyoo}/\textit{an}, go/row/line)
require context-dependent disambiguation.
Moreover, \begin{CJK}{UTF8}{ipxm}方\end{CJK} (way/person) is read \textit{hoo} when indicating direction or comparison, but \textit{kata} when referring to a person.

\textbf{Pronunciation rules.}
Certain kana characters change pronunciation in specific grammatical contexts.
The particles \begin{CJK}{UTF8}{ipxm}は\end{CJK}(\textit{ha}), \begin{CJK}{UTF8}{ipxm}を\end{CJK}(\textit{wo}), and \begin{CJK}{UTF8}{ipxm}へ\end{CJK}(\textit{he})
are pronounced \textit{wa}, \textit{o}, and \textit{e}, respectively.
Consecutive vowels in Japanese are often realized as long vowels.
For example, \begin{CJK}{UTF8}{ipxm}お父さん\end{CJK} (\textit{otousan}, father) is pronounced \textit{otoosan},
where the \textit{ou} sequence becomes a long vowel \textit{oo}.

\textbf{Numeral--counter expressions.}
Numeral--counter combinations follow complex phonological rules
involving sequential voicing and gemination.
For example, \begin{CJK}{UTF8}{ipxm}1本\end{CJK} (\textit{ichi}+\textit{hon}, one long object) is pronounced \textit{ippon},
and \begin{CJK}{UTF8}{ipxm}2人\end{CJK} (\textit{ni}+\textit{nin}, two people) is pronounced \textit{futari}.
These irregular changes depend on both the numeral and the counter,
making rule-based handling difficult.


\begin{figure}[t]
  \centering
  \includegraphics[width=0.95\linewidth]{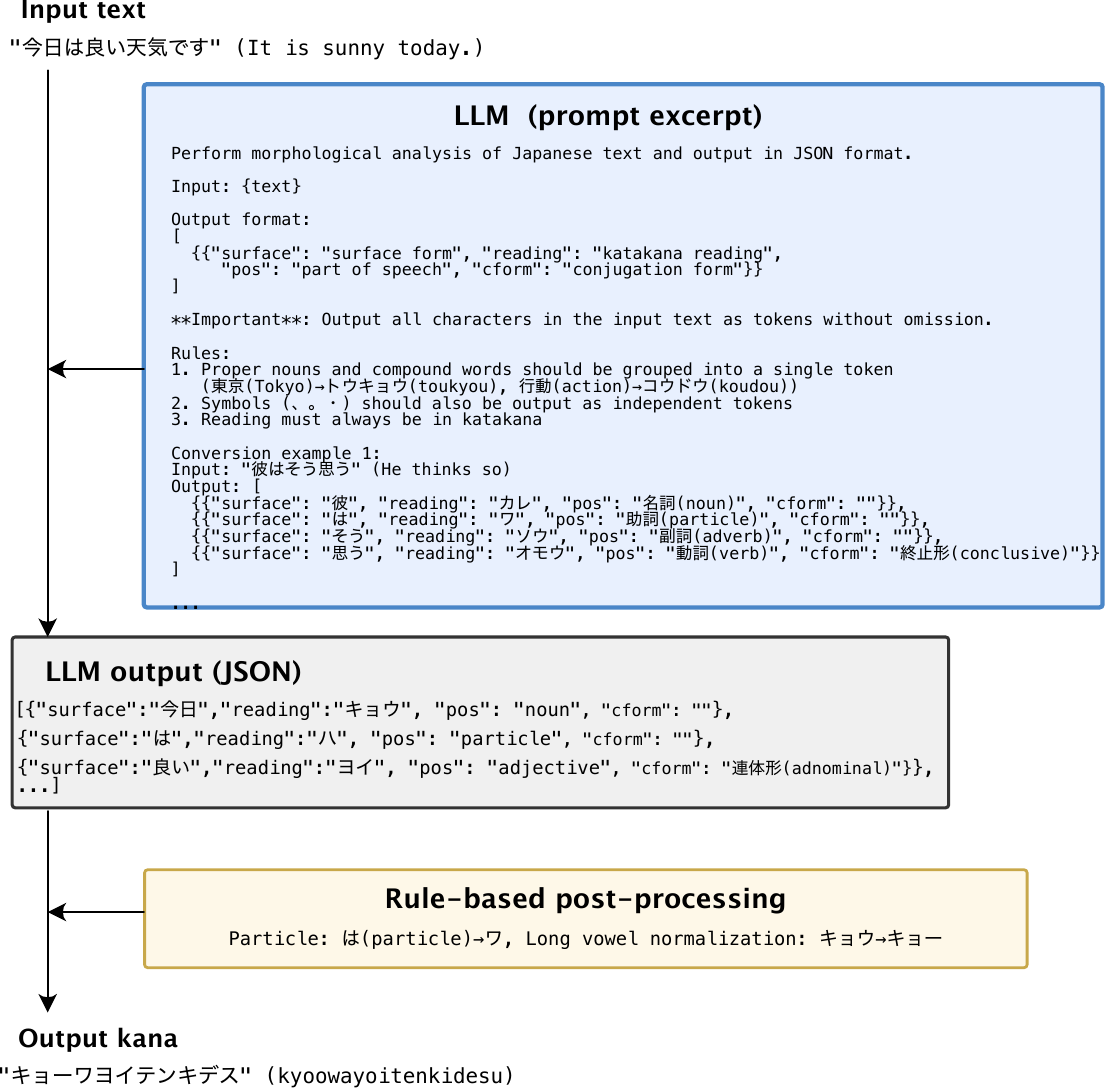}
  \caption{Parse mode pipeline. The LLM performs morphological analysis and outputs JSON with word readings. Rule-based post-processing applies pronunciation rules.}
  \label{fig:parse_mode}
\end{figure}

\begin{figure}[t]
  \centering
  \includegraphics[width=0.95\linewidth]{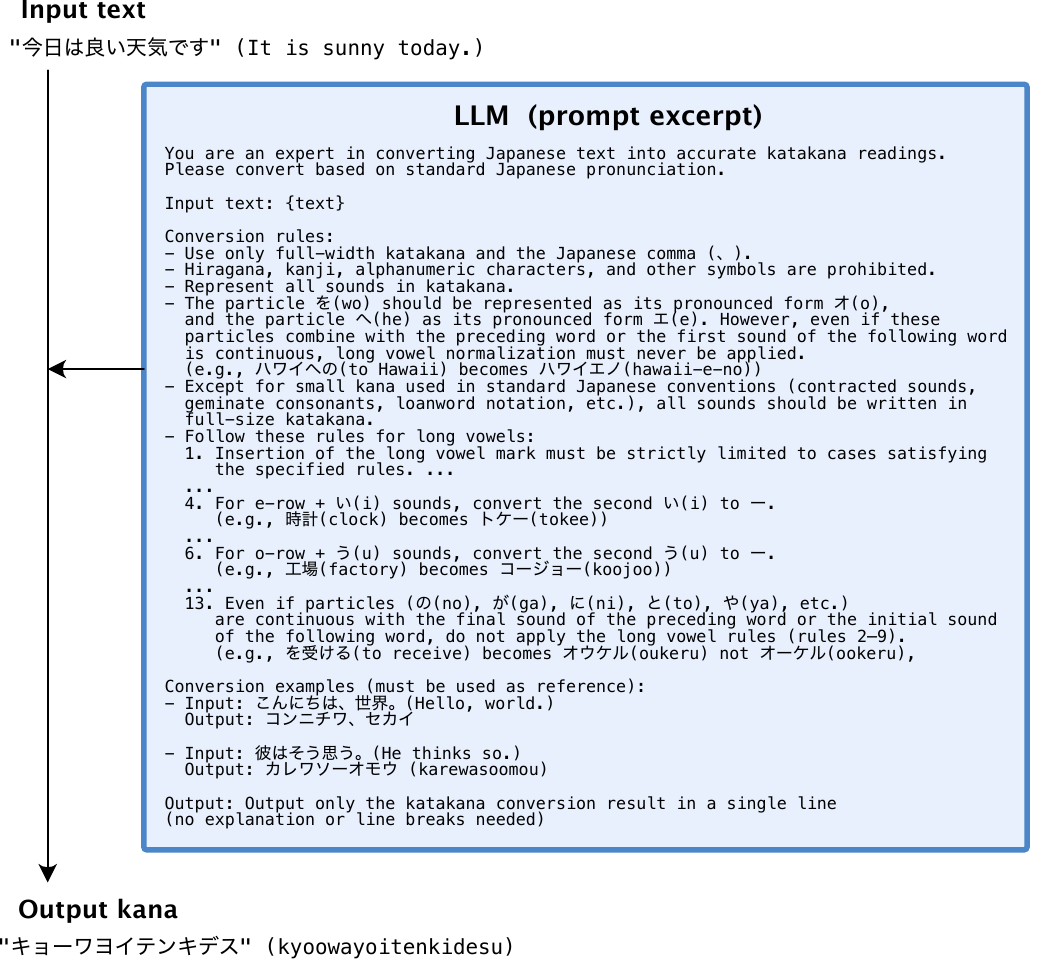}
  \caption{Direct mode pipeline. The LLM directly converts the input text to a kana reading in a single step.}
  \label{fig:direct_mode}
\end{figure}

\section{LLM-based G2P}

We evaluate two approaches for using LLMs in Japanese G2P,
inspired by the cascade and direct methods studied by Fetrat Qharabagh et al.~\cite{mahta2024llm}.
Ideally, the LLM would directly convert text to kana in a single step, which we refer to as direct mode.
However, our preliminary experiments showed that
instructing the LLM to handle all pronunciation rules such as particle conversion and vowel lengthening through prompting
makes the instructions complex and increases errors.
We therefore also propose parse mode,
where rule-based post-processing handles deterministic pronunciation rules,
keeping the LLM's prompt simple and focused on word segmentation and reading estimation.

\subsection{Parse mode}

In parse mode, an LLM performs morphological analysis:
given an input sentence, it outputs a sequence of words with their kana readings.
The LLM replaces the morphological analyzer (e.g., MeCab~\cite{kudo2004mecab}) in a conventional G2P pipeline.
A rule-based post-processing step then applies pronunciation rules
such as particle conversion
(\begin{CJK}{UTF8}{ipxm}は\end{CJK}(\textit{ha}) $\rightarrow$ \begin{CJK}{UTF8}{ipxm}ワ\end{CJK}(\textit{wa}))
and long vowel normalization,
producing the final kana sequence.
Since the rule-based step handles pronunciation rules deterministically,
the LLM only needs to perform word segmentation and reading estimation.
Fig.~\ref{fig:parse_mode} illustrates the parse mode pipeline.

The prompt instructs the LLM to output a JSON array
where each element contains the surface form, katakana reading, and part-of-speech of a word.
Compound nouns and proper nouns are grouped as single tokens.
Few-shot examples are provided to demonstrate the expected output format.

\subsection{Direct mode}

In direct mode, the LLM directly outputs the full kana reading of the input sentence
without intermediate morphological analysis.
This approach places the entire G2P burden on the LLM,
including word segmentation, reading estimation, and pronunciation rule application.
While simpler in pipeline design, it requires the LLM to handle all Japanese pronunciation rules.
Fig.~\ref{fig:direct_mode} illustrates the direct mode pipeline.

The prompt specifies detailed rules for katakana conversion,
including particle pronunciation, long vowel normalization for each vowel class
(e.g., \textit{o}+\begin{CJK}{UTF8}{ipxm}う\end{CJK}(\textit{u}) $\rightarrow$ long vowel (\textit{oo})),
and exceptions such as verb inflectional endings and particle boundaries
where lengthening must not apply.
In addition, few-shot examples are provided.

\section{Experiments}

\subsection{Dataset and evaluation metric}

We used 3,000 sentences from the nonpara30 subset
of the JVS (Japanese versatile speech) corpus~\cite{takamichi2020jvs}.
The sentences cover diverse phenomena, including onomatopoeia and loanwords,
which are frequently out-of-vocabulary
and thus particularly challenging for conventional dictionary-based tools.
Using UniDic-based morphological analysis,
we found proper nouns written in kanji in 181 sentences (6.0\%)
and those written in katakana in 256 sentences (8.5\%),
although this automatic detection does not capture all such cases.
Numerals, whose reading depends on the following counter,
appear in 14.2\% of the sentences.
We manually annotated kana readings for all sentences as reference transcriptions.
The kana character error rate (CER) was computed between predicted and reference kana sequences.
Since Japanese kana have a nearly one-to-one correspondence with phonemes,
kana CER can be regarded as essentially equivalent to phoneme error rate.
Before computing CER, punctuation marks were removed
and long vowels were normalized to account for acceptable variation in kana representation.

\subsection{Evaluated models}

We evaluated three categories of models.
\textbf{Proprietary LLMs} included
Claude Opus~4.6 and Sonnet~4.6 (Anthropic),
Gemini~3.1~Pro, 3~Flash, and 2.5~Flash (Google),
and GPT-5.2 (OpenAI).
\textbf{Open-weight LLMs} included
models from the Gemma, Qwen, Llama, GLM, gpt-oss, and Kimi families,
ranging from 2B to 1T parameters.
Some models were Japanese-specialized through continual pretraining
(denoted ``Swallow''~\cite{fujii2024swallow}).
Most open-weight models were deployed locally;
larger models (GLM-4.7, GLM-5, Kimi~K2.5, Qwen3.5-397B) were accessed via their providers' APIs
(marked with $\dagger$ in Table~\ref{tab:main_results}).
\textbf{Conventional morphological analyzers} included
OpenJTalk~\cite{openjtalk}, MeCab+IPAdic and MeCab+UniDic~\cite{kudo2004mecab},
KyTea~\cite{neubig2011kytea}, KWJA~\cite{kwja}, Sudachi~\cite{takaoka2018sudachi}, and Vaporetto~\cite{akabe2024vaporetto}.
All LLM results in Table~\ref{tab:main_results} were obtained
with reasoning disabled (non-thinking mode or reasoning effort set to low).
For conventional tools, the same rule-based post-processing as in parse mode
(particle conversion and long vowel normalization) was applied to their output.
Since both share this identical post-processing and differ only in the morphological
analysis step (an LLM vs. a conventional analyzer),
the gap between them directly reflects the difference in analysis capability.

\begin{table}[t]
  \caption{Kana CER (\%) for all evaluated models. The best value in each column is in bold. $\dagger$: open-weight models accessed via API.}
  \scriptsize
  \setlength{\tabcolsep}{4pt}
  \label{tab:main_results}
  \centering
  \begin{tabular}{ l c r r }
    \toprule
    Model & Size & Parse & Direct \\
    \midrule
    \multicolumn{4}{l}{\textit{Proprietary LLM}} \\
    Claude Opus 4.6         & --              & \textbf{0.52} & 0.74  \\
    Claude Sonnet 4.6       & --              & 0.82          & 1.28  \\
    Gemini 2.5 Flash        & --              & 1.08          & 1.17  \\
    Gemini 3 Flash          & --              & 0.94          & 1.61  \\
    Gemini 3.1 Pro          & --              & 0.62          & \textbf{0.53} \\
    OpenAI GPT-5.2          & --              & 1.64          & 1.05  \\
    \midrule
    \multicolumn{4}{l}{\textit{Open-weight LLM}} \\
    CALM3-22B               & 22B             & 14.30 & 22.81 \\
    Gemma3-4B               & 4B              & 34.82 & 56.69 \\
    Gemma2-27B              & 27B             & 10.33 & 24.56 \\
    Gemma2-Swallow-27B      & 27B             & 5.74  & 17.53 \\
    Gemma3-12B              & 12B             & 14.15 & 28.77 \\
    Gemma3-27B              & 27B             & 5.75  & 15.16 \\
    GLM4-9B                 & 9B              & 26.89 & 57.56 \\
    GLM4-32B                & 32B             & 16.82 & 35.62 \\
    GLM4.7$^\dagger$       & 355B            & 3.70  & 5.87  \\
    GLM5$^\dagger$          & 744B            & 1.44  & 4.96  \\
    gpt-oss-20b             & 20B             & 7.95  & 14.71 \\
    gpt-oss-120b            & 120B            & 3.03  & 3.45  \\
    Kimi K2.5$^\dagger$      & 1T      & 1.35  & 4.91  \\
    Llama3.1-Swallow-8B     & 8B              & 9.34  & 22.47 \\
    Llama3.3-70B            & 70B             & 6.58  & 23.48 \\
    Llama3.3-Swallow-70B    & 70B             & 2.85  & 11.78 \\
    llm-jp-3.1-13b          & 13B             & 15.36 & 86.02 \\
    Qwen2.5-7B              & 7B              & 43.27 & 79.39 \\
    Qwen2.5-32B             & 32B             & 16.54 & 35.48 \\
    Qwen3-4B                & 4B              & 35.77 & 57.72 \\
    Qwen3-8B                & 8B              & 28.13 & 100.15 \\
    Qwen3-14B               & 14B             & 17.13 & 32.11 \\
    Qwen3-32B               & 32B             & 17.07 & 32.46 \\
    Qwen3-Swallow-32B       & 32B             & 9.30  & 21.59 \\
    Qwen3.5-2B              & 2B              & 41.87 & 87.30 \\
    Qwen3.5-4B              & 4B              & 23.00 & 43.98 \\
    Qwen3.5-9B              & 9B              & 14.17 & 27.47 \\
    Qwen3.5-27B             & 27B             & 6.27  & 14.61 \\
    Qwen3.5-35B-A3B       & 35B             & 6.12  & 26.93 \\
    Qwen3.5-122B-A10B$^\dagger$    & 122B            & 7.57  & 16.80 \\
    Qwen3.5-397B$^\dagger$   & 397B            & 1.96  & 6.20  \\
    \midrule
    \multicolumn{4}{l}{\textit{Conventional morphological analyzer}} \\
    KWJA                    & --  & 2.52 & -- \\
    KyTea                   & --  & 1.42 & -- \\
    MeCab+IPAdic            & --  & 1.78 & -- \\
    MeCab+UniDic            & --  & 1.54 & -- \\
    OpenJTalk               & --  & 1.03 & -- \\
    Sudachi                 & --  & 1.83 & -- \\
    Vaporetto               & --  & 1.86 & -- \\
    \bottomrule
  \end{tabular}
\end{table}

\begin{figure}[t]
  \centering
  \includegraphics[width=\linewidth]{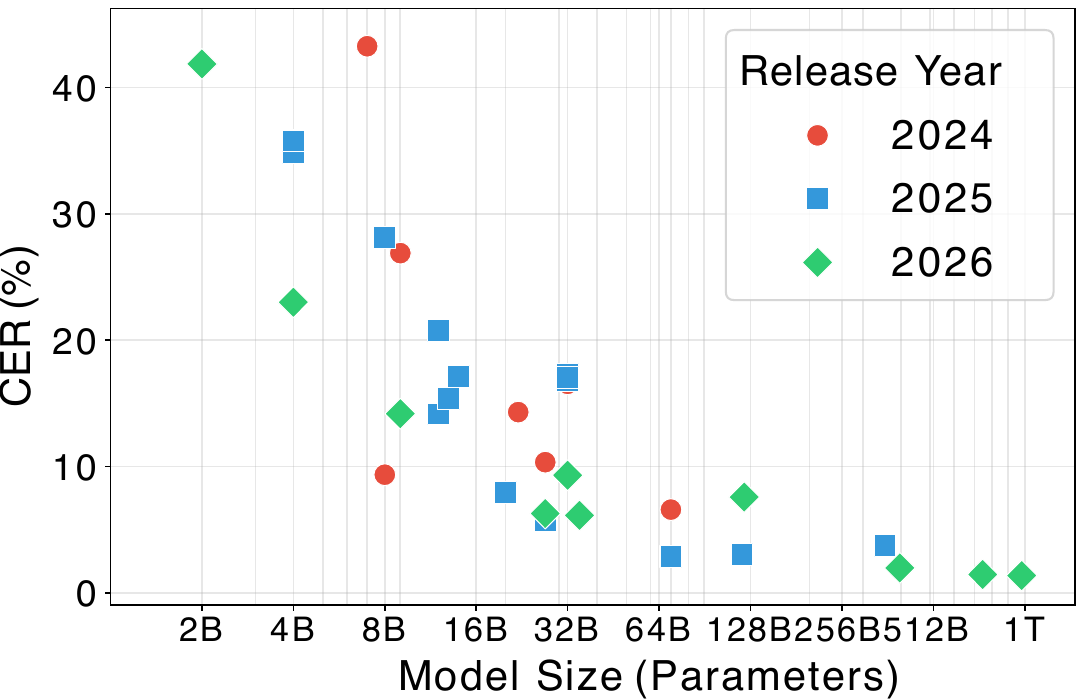}
  \caption{Model size vs. kana CER (\%) in parse mode for open-weight LLMs.}
  \label{fig:size_vs_cer}
\end{figure}

\subsection{Results}

Table~\ref{tab:main_results} shows kana CER for all evaluated models.

\textbf{Effect of model size.}
Fig.~\ref{fig:size_vs_cer} shows model size vs. CER in parse mode for local LLMs.
Larger models consistently achieved lower CER within the same model family:
Gemma3 improved from 34.82\% (4B) to 14.15\% (12B) to 5.75\% (27B),
Qwen2.5 from 43.27\% (7B) to 16.54\% (32B),
and Qwen3.5 from 41.87\% (2B) to 23.00\% (4B) to 14.17\% (9B) to 6.27\% (27B).
This indicates that the scaling law~\cite{kaplan2020scaling} also holds for this G2P task,
with CER decreasing monotonically as model size increases.
Smaller models exhibited three types of failures:
(1) substituting unrelated words rather than providing faithful readings
(e.g., Gemma3-4B converted
\begin{CJK}{UTF8}{ipxm}人と出会う\end{CJK} (\textit{hito to deau}, to meet people) to \textit{\textbf{tojin to dooga}}),
(2) producing incorrect kanji readings
(e.g., \begin{CJK}{UTF8}{ipxm}生涯\end{CJK} (\textit{shoogai}, lifetime) as \textit{\textbf{shingai}}),
and (3) failing to preserve even hiragana characters
(e.g., \begin{CJK}{UTF8}{ipxm}すべて\end{CJK} (\textit{subete}, all) as \textit{\textbf{sabute}}).
Newer model versions also improved CER at similar sizes:
Gemma3-27B (5.75\%) outperformed Gemma2-27B (10.33\%),
and Qwen3.5-27B (6.27\%) outperformed both Qwen2.5-32B (16.54\%) and Qwen3-32B (17.07\%).
API models achieved the lowest CER overall:
Claude Opus~4.6 in parse mode achieved 0.52\% and
Gemini~3.1~Pro in direct mode achieved 0.53\%,
both outperforming the best conventional tool, OpenJTalk (1.03\%).

\textbf{Effect of Japanese-specialized training.}
Swallow variants, which undergo continual pretraining on Japanese data,
consistently outperformed their base models:
Llama3.3-Swallow-70B (2.85\%) vs. Llama3.3-70B (6.58\%),
Qwen3-Swallow-32B (9.30\%) vs. Qwen3-32B (17.07\%),
and Gemma2-Swallow-27B (5.74\%) vs. Gemma2-27B (10.33\%).
These results suggest that Japanese-specialized training effectively reduced CER.
Llama3.3-Swallow-70B achieved the best CER among all locally deployed LLMs.

\textbf{Parse mode vs. direct mode.}
For all local LLMs and most API models, parse mode yielded lower CER than direct mode.
The advantage was particularly large for smaller models:
Gemma3-4B showed 34.82\% in parse mode vs. 56.69\% in direct mode.
This is because direct mode had to handle particle conversion and long vowel normalization within the LLM itself,
which many models struggled with.
For example, given \begin{CJK}{UTF8}{ipxm}母の死は私の生涯に大きな空白を残した\end{CJK} (My mother's death left a great void in my life),
Llama3.3-Swallow-70B in parse mode correctly produced the reading,
while in direct mode it output \textit{haha no shi \textbf{ha} \ldots\ sh\textbf{ou}gai \ldots\ kuuhaku \textbf{wo} nokoshita},
failing to convert the particles
\begin{CJK}{UTF8}{ipxm}は\end{CJK}(\textbf{ha})$ \rightarrow$ \textit{wa} and
\begin{CJK}{UTF8}{ipxm}を\end{CJK}(\textbf{wo})$ \rightarrow$ \textit{o},
and to normalize the long vowel \textit{ou}$ \rightarrow$ \textit{oo}.
For smaller local LLMs, direct mode sometimes failed to follow the prompt instructions entirely,
resulting in extremely high CER for models such as llm-jp-3.1-13b (86.02\%) and Qwen3-8B (100.15\%).

Notable exceptions were Gemini~3.1~Pro and GPT-5.2, where direct mode was slightly better.
For these models, direct mode successfully followed the prompt instructions
and rarely produced particle conversion or long vowel errors.
In addition, parse mode introduced errors by splitting numeral--counter expressions during morphological analysis:
for example, \begin{CJK}{UTF8}{ipxm}2人\end{CJK}(two people) was segmented into
\begin{CJK}{UTF8}{ipxm}2\end{CJK} and \begin{CJK}{UTF8}{ipxm}人\end{CJK},
producing \textit{ni-nin} instead of the correct compound reading \textit{futari}.
Direct mode, which processes the text without segmentation, correctly handled such cases.

\textbf{LLMs vs. conventional tools.}
Claude Opus~4.6 (0.52\%) and Gemini~3.1~Pro (0.62\%) clearly outperformed
the best conventional tool, OpenJTalk (1.03\%), in parse mode.
LLMs were stronger on out-of-vocabulary words that are absent from the dictionary of conventional tools:
for example, OpenJTalk misread
\begin{CJK}{UTF8}{ipxm}剣歯虎\end{CJK}(saber-toothed tiger) as \textit{ken\textbf{patora}} instead of \textit{kenshiko},
\begin{CJK}{UTF8}{ipxm}海の幸\end{CJK}(bounty) as \textit{umi no \textbf{koo}} instead of \textit{umi no sachi},
and \begin{CJK}{UTF8}{ipxm}甘味料\end{CJK}(sweetener) as \textit{\textbf{ama}miryoo} instead of \textit{kanmiryoo}.
Conversely, LLMs as probabilistic models occasionally produced unlikely readings for common words:
Claude Opus misread
\begin{CJK}{UTF8}{ipxm}本名\end{CJK}(real name) as \textit{\textbf{honme}} instead of \textit{honmyoo}.

\textbf{Effect of thinking (reasoning) mode.}
We compared thinking \cite{deepseek2025r1} and non-thinking modes for three models in parse mode.
For Qwen3-32B (non-thinking 17.07\% vs. thinking 17.32\%) and gpt-oss-20b (non-thinking 7.95\% vs. thinking 7.91\%),
the differences were negligible.
However, Gemini~3~Flash showed a clear improvement from non-thinking 0.94\% to thinking 0.54\%,
suggesting that reasoning can further improve accuracy for already high-performing models.
A detailed analysis revealed that the thinking mode produced more accurate word segmentation,
which in turn allowed the long vowel normalization rule to be applied correctly,
since vowel lengthening must not apply across word boundaries.

\section{Discussion: Comparison with E2E TTS}

To investigate the effectiveness of G2P on TTS,
we compared the pronunciation accuracy of G2P-based TTS and E2E TTS systems.

For the experiments,
we fine-tuned CosyVoice~2~\cite{du2024cosyvoice2} with LoRA~\cite{hu2022lora}
on the Corpus of Spontaneous Japanese (CSJ)~\cite{Maekawa_2003}
to accept kana input.
For G2P-based synthesis, LLM-predicted kana sequences were fed into this kana-input TTS.
We compared against E2E TTS systems that accept raw text:
CosyVoice~2, Gemini~2.5~Flash~TTS, Qwen~3~TTS, and ElevenLabs~v2.
Pronunciation accuracy was evaluated by recognizing synthesized speech
with a kana-output ASR model (Whisper~\cite{radford2023robust} fine-tuned for kana output,
achieving 2.22\% CER on the CSJ test set)
and computing CER against the reference kana.
Estimated speech naturalness was measured by UTMOS~\cite{saeki2022utmos}.

\begin{table}[t]
  \caption{Pronunciation accuracy (CER \%) and estimated naturalness (UTMOS) of G2P-based and E2E TTS systems.}
  \footnotesize
  \label{tab:tts_comparison}
  \centering
  \begin{tabular}{ l l r r }
    \toprule
    System & Input & CER & UTMOS \\
    \midrule
    \multicolumn{4}{l}{\textit{G2P + Kana-input TTS (fine-tuned CosyVoice 2)}} \\
    True kana (oracle)       & Kana & 2.10 & 3.81 \\
    Gemini 3.1 Pro (direct)  & Kana & 2.38 & 3.82 \\
    Claude Opus 4.6 (parse)  & Kana & 2.69 & 3.82 \\
    \midrule
    \multicolumn{4}{l}{\textit{E2E TTS}} \\
    Gemini 2.5 Flash TTS     & Text & 3.96 & 3.75 \\
    Qwen 3 TTS (1.7B)        & Text & 4.31 & 4.00 \\
    CosyVoice 2              & Text & 12.08 & 3.45 \\
    ElevenLabs v2             & Text & 13.96 & 3.36 \\
    \bottomrule
  \end{tabular}
\end{table}

Table~\ref{tab:tts_comparison} shows the results.
G2P-based TTS using Gemini~3.1~Pro (direct mode) achieved 2.38\% CER,
which is close to the oracle condition using true kana (2.10\%)
and much lower than all E2E TTS systems.
Even Gemini~2.5~Flash~TTS, one of the state-of-the-art E2E systems,
showed 3.96\% CER, higher than the G2P-based systems.
UTMOS scores for G2P-based TTS (3.82) were comparable to
those of E2E systems (3.36--4.00),
indicating that the explicit G2P stage did not degrade speech naturalness.
These results confirmed that an explicit G2P module
provided a practical advantage in pronunciation accuracy
while maintaining naturalness.

\section{Conclusions}

We presented a large-scale benchmark of LLM-based G2P conversion for Japanese.
The best proprietary API models achieved kana CER below 0.6\%,
outperforming conventional morphological analyzers.
Parse mode was more effective than direct mode for most models,
and Japanese-specialized training greatly improved local LLM performance.
We also demonstrated that combining LLM-based G2P with kana-input TTS
achieved better pronunciation accuracy than E2E TTS while maintaining comparable naturalness.

For future work,
tuning prompts based on LLM output patterns, combining multiple models,
or replacing rule-based post-processing with LLM-based processing may further reduce errors.
Evaluating G2P on proper nouns is also important,
as these are a major source of errors in practical TTS.
Moreover, in-context learning could let users specify their readings via prompting,
a unique advantage of LLM-based G2P over conventional tools.

\section{Generative AI Use Disclosure}
Claude Code, ChatGPT, and Gemini were used for manuscript editing.

\bibliographystyle{IEEEtran}
\bibliography{mybib}

\end{document}